# LLMs for Academic Workflows: An Evaluation of Literature Reviews Generated with Short and Long Context Windows of LLMs

**Muhammad Ali Chaudhry**[1,2], **Xinyuan Hao**[1], **Haifa Alwahaby** [1]

Ali@ResearchPal.co, Xinyuan.Hao.21@ucl.ac.uk, Haifa.Alwahaby.18@ucl.ac.uk

[1] University College London, [2] ResearchPal

## Abstract

Our research focuses on evaluating literature reviews generated in short and long context settings of large language models (LLMs) to investigate the impact of context window on the quality of AI-generated literature reviews and the role of AI in supporting literature review writing. Twenty AI-generated literature reviews based on research sources from Semantic Scholar and Arxiv were evaluated by two researchers across 15 dimensions. Our findings reveal that AI-generated literature reviews require human oversight to meet academic publishing standards. As context windows increase, LLMs can incorporate broader information and maintain coherence across longer inputs, but they also exacerbate issues such as content repetition, omission of critical work, and a tendency towards descriptiveness over synthesis. Our work shows that AI-generated reviews can provide foundational overviews, but their output must be critically evaluated and refined by domain experts. Future research should consider integrating other LLMs and fine-tuned models in different domains with hybrid approaches that combine human expertise with AI capabilities to address the limitations identified in this study.

## 1. Introduction

The rapid evolution of Large Language Models (LLMs) has catalysed transformative advancements in natural language processing (NLP). Central to the efficacy of these models is their ability to process and generate text within a "context window," which defines the amount of sequential input the model can consider during a single inference (Chen et al, 2023, Fei et al., 2023, Chang et al, 2023). The size of this context window plays a pivotal role in determining the model's capacity to understand, maintain coherence, and generate responses informed by extended textual inputs (Zhang et al., 2024, Hosseini et al., 2024, Zhu et al, 2023). Recent efforts in research and development of LLMs have prioritised expanding context windows, driven by the increasing demand for models capable of handling tasks that require comprehension and synthesis of long-form content, such as document summarisation, multi-document analysis or detailed detailed review of enterprise text data. Traditionally, the computational and architectural limitations of transformer-based LLMs (Huang et al, 2023, Peng et al, 2024) constrained the feasible size of context windows. The quadratic scaling of attention mechanisms posed significant barriers to extending the context length without incurring prohibitive computational costs. However, recent breakthroughs have introduced innovative methodologies, including sparse attention mechanisms (Chen et al, 2023), memory-efficient architectures (Zhao et al., 2024), and external memory integration (Andriopoulos and Pouwelse, 2023). These advancements have not only mitigated the computational challenges but have also redefined the potential applications of LLMs by enabling them to perform effectively on tasks that demand long-range dependencies and intricate contextual understanding.

This expansion of context windows in LLMs has had a profound impact on their applications, significantly broadening their utility and effectiveness in various domains. This includes improved long-form text generation and summarisation for academic literature reviews, enhanced understanding of long documents such as books and academic research papers, ability to process multiple documents simultaneously, better data retrieval and knowledge integration from diverse sources, and significantly more time efficiency in processing large amounts of text.

We used literature reviews generated from long context window rather than more traditional approaches such as retrieval augmented generation (RAG) (Fan et al, 2024; Gao et al, 2023). Li et al (2024) have shown that long context generally outperforms RAG for question answering tasks from documents.

This research investigates the usability of LLMs on the quality of AI generated academic literature reviews and identifies the role and responsibilities for human-in-the-loop (researchers, students and industry professionals) for reviewing the AI generated literature reviews before using them.

## 2. Literature Review

Conducting an academic literature review involves systematically identifying a research gap, defining a research question to address that particular gap, searching for relevant academic literature and research sources, critically evaluating and selecting credible sources, organizing the research sources thematically or chronologically, analyzing and synthesizing the information to identify patterns and gaps, writing the review with a clear structure, and properly citing all sources to acknowledge original authors and facilitate further research (Figure 1).

The integration of LLMs into academic workflows such as literature reviews has sparked significant interest in the research community. LLMs have been explored as tools for automating and augmenting the literature review process. They excel at tasks such as text summarization, semantic search, and thematic synthesis, which are integral to systematic literature reviews. Research by Longston and Ashford (2024) highlights the ability of LLMs to summarise abstracts from multiple documents. Si et al (2024) demonstrated that LLMs could provide thematic clustering of research articles, streamlining the identification of trends and gaps in literature, potentially leading to novel research ideas. Liam et al (2024) mapped the use of LLMs in scientific papers. Their large-scale systematic review revealed an increasing use of LLMs for academic writing, particularly in computer science papers. Alex Zhang's (2024) research identified a change in word frequencies in academic publications since the launch of Chatgpt.

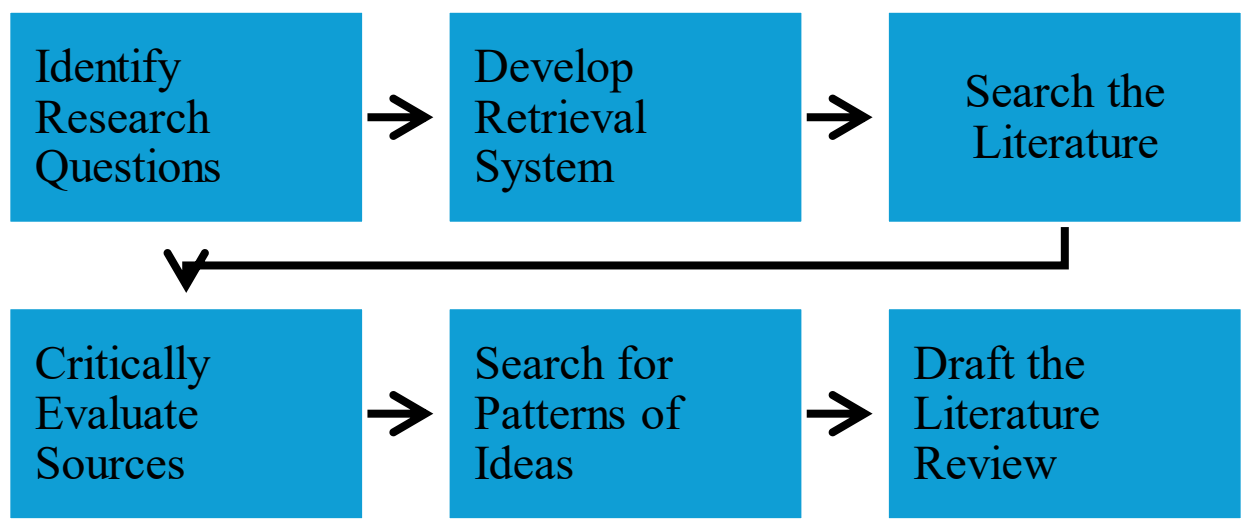

*Figure 1.* Steps to producing a literature review

Tyser et al (2024) used a different approach to address some of the limitations of using LLMs as reviewers and evaluators, thereby enhancing the quality of the reviewing process. They have presented a demo for an LLM reviewing system that consistently delivers high-quality reviews. Their work mitigates the risks of misuse of LLMs in the review process including inflated review scores, overconfident ratings, and skewed score distributions by augmenting the LLM with multiple documents such as the review form, reviewer guide, code of ethics and conduct.

Finding relevant research papers from the academic databases is a key component of good quality literature reviews. The integration of semantic search functionality in LLMs enables them to identify relevant papers based on user-defined queries or prompts. Unlike traditional keyword-based search methods, semantic search leverages the contextual understanding of LLMs to retrieve papers that align closely with the intended meaning, as shown by Yang et al. (2022). This has proven particularly useful for interdisciplinary research, where terminologies and concepts vary across domains.

Jung et al (2024) found significant differences in the trust levels of researchers for using LLMs. They found that it is strongly correlated with their usage time and frequency. While there has been an extensive use of LLMs in academia and research through AI-powered tools such as ResearchPal[1], the companies building these tools strongly encourage human oversight before final publications.

## 3. Methodology

In order to assess the impact of a long context window on academic literature reviews, twenty literatures were generated with 10 different research questions related to AI in education. We generated ten literatures with a long context window and ten with a short context window.

The literature was rated by two researchers who were experts in AI for Education. There were five literatures given to each researcher with two versions a and b of the same research question, so that each researcher could compare the two versions, with a total of 10 literatures per researcher and 20 literatures in total.

A parallel selection process was conducted on 10% of the literature to determine the level of inter-rater agreement. Due to the ordinal nature of the data, which consists of ratings between 1 and 5, weighted Cohen's Kappa is used. Resulting in a Weighted Cohen's Kappa = 0.743, which indicates substantial agreement (McHugh, 2012). Disagreements between the two researchers were resolved through meetings and discussions. This resulted in further clarification of the rating criteria.

All the literature reviews were evaluated across 15 dimensions shown in Appendix 1. Each category is rated from 1 to 5, with 75 being the highest total.

All the papers were extracted from two open access academic databases: semantic scholar and arxiv. For each query, the top hundred papers were extracted from semantic scholar and the top fifty papers were selected from arxiv. Gemini 1.5 pro was used to write the literature review in two different formats: long context, and short context with three iterations. Iterations were used with the short context length to bring the length of literature review at par with the long context window. Some experimentation with the prompts was done to improve the quality of literature reviews generated. Eventually one prompt was used for both formats to ensure consistency of responses with the only differentiating factor being the context window.

[1] https://researchpal.co/

The rubric was developed by taking into consideration various literature on how to conduct a literature review. First and foremost, a well-done literature review should cover each of the major themes and subthemes found within the general research topic. By using subtopics of the general research question, the literature review progresses from a general overview to a more specific focus (Denney & Tewksbury, 2013). As a result, the first dimension of the rubric focused on the structure and organization of the literature review. Maintaining a logical organization with clear transitions between sections and ideas. Moreover, an effective literature review is characterized by the author's attempt to critically evaluate and examine the relevant literature (Chigbu et al., 2023). Therefore, it was important for analysis criteria to be included as an evaluation criterion in the rubric to ensure that critical thinking is present and that evaluations of studies are indicated. Literature reviews often end up being merely descriptive summaries, only providing an overview of topics (Snyder, 2024). In response to this common limitation, the rubric included synthesis criteria, which ensures that articles support ideas rather than simply summarizing them. It is important to understand that a comprehensive literature review includes a critical evaluation of the content or material relevant to a specific subject or topic in order to identify any gaps in the literature and address them. The rubric included the identification of research gaps within the literature review as a concluding point within the conclusion criteria (Chigbu et al., 2023). With the vast computing capabilities of LLMs such as ChatGPT, an enormous amount of information may be generated, some of which may not be accurate or relevant. It was therefore important to include accuracy and relevance as evaluation metrics in the rubric (Antu et al., 2023). It was therefore important to include content's accuracy and relevance as evaluation metrics in the rubric.
When it comes to citation quality, articles published in academic journals and books are the most appropriate sources of evidence (Denney & Tewksbury, 2013). It is due to the peer-review process that papers published in journals and books undergo. Accordingly, peer-reviewed sources were considered as criteria for evaluating the quality of a source. Based on the fact that the evaluated literature was cited according to the American Psychological Association style, assessment of the correctness of formatting was an integral part of the rubric. The currency of the citation was determined by considering the majority of references, not necessarily the full number of references, as historical references should be considered in some instances.

After literature reviews were scored by the researchers, total scores were categorised in four groups: excellent, good, pass and fail. The scoring rubric is shown in appendix 1 and scoring details for each group are shown in table 1 below.

*Table.1*: Scoring categorisation for the rubric

| Category | Mark Range (Out of 75) |
|---|---|
| **Excellent** | Lit review > 60 |
| **Good** | 45 < Lit review ≤ 60 |
| **Pass** | 30 < Lit review ≤ 45 |
| **Fail** | Lit review ≤ 30 |

## 4. Findings

All the AI-generated literature reviews managed to reach at least the good level, which demonstrated great potential of using AI as a supporting tool to write literature reviews. Table 2 and 3 show the scores for all the literature reviews.

Even though both long context window LLM and short-context window LLM performed well in generating literature reviews, we also observed several limitations and some differences in the reviews that are potentially related to the context window size.

*Table.2:* Scores for all the literature reviews graded by Rater 1

| Literature Review | Rater 1 Mark | Category |
|---|---|---|
| 1A | 58 | Good |
| 1B | 59 | Good |
| 2A | 53 | Good |
| 2B | 53 | Good |
| 3A | 59 | Good |
| 3B | 58 | Good |
| 4A | 61 | Excellent |
| 4B | 53 | Good |
| 5A | 59 | Good |
| 5B | 52 | Good |

### 4.1 Repetition of Content

A recurring issue in AI-generated literature reviews is the repetition of themes, concepts, and even verbatim content. This redundancy is especially prevalent in reviews produced with larger context windows (Category A). For instance, many sections within these reviews tend to repeat the same arguments in different expressions, thus undermining coherence and depth of the literature reviews. Although reviews generated with smaller context windows (Category B) also presented the same issue, the frequency and extent of repetition are noticeably lower, suggesting that larger context capacities may inadvertently increase redundancy.
However, it is also worth mentioning that reviews created by larger context window LLMs tended to cite more relevant previous literature, which might boost the reliability of the reviews, especially for novice learners in the fields. This finding indicates that a larger context window might help improve reliability of the reviews by processing and citing more literature at once, but there might be a trade-off as it may also struggle to effectively synthesize this input into non-repetitive, cohesive narratives.

> *This question is at the forefront of educational research, as the potential of AI to tailor learning to individual needs is immense……Literature Review: How can AI personalize learning experiences for individual students? This question is at the forefront of educational research, as the potential of AI to tailor learning to individual needs is immense. – from LR 1B*

*Table 3*: Scores for all the literature reviews graded by Rater 2

| Literature Review | Rater 2 Mark | Category |
|---|---|---|
| 6A | 52 | Good |
| 6B | 55 | Good |
| 7A | 52 | Good |
| 7B | 56 | Good |
| 8A | 59 | Good |
| 8B | 61 | Excellent |
| 9A | 57 | Good |
| 9B | 54 | Good |
| 10A | 59 | Good |
| 10B | 53 | Good |

**4.2 Omission of Key Works**

It is intriguing to notice that some AI-powered literature reviews, regardless of the context window size, fail to include essential works in their respective fields. For example, a review about ethics of AI and algorithmic bias in educational settings (LR 9A) failed to reference any works by Dr. Wayne Holmes or Dr. Ryan Baker, which are critical literature with high citations. This limitation indicates that the models might encounter difficulties in comprehensively mapping the breadth of domain knowledge.

The researchers are aware that this omission might relate to the limited open-source archives used as inputs for the LLMs. However, the issue still raises questions about the ability of the LLMs to present a complete and accurate state-of-the-art summary of a topic.

**4.3 Descriptive Nature**

Many AI-generated literature reviews appeared to be descriptive summaries of the previous works rather than analytical synthesis or critical evaluations. This trend is particularly obvious in longer reviews (Category A), where the increased length tended to be achieved by mentioning more prior articles without offering substantive critical evaluations. The descriptive nature of the reviews showcased the disappointing ability of AI in making meaningful connections between sources or contributing innovative arguments. This finding diminishes the overall scholarly value of the AI-generated literature reviews and proves the limited ability of AI as a fully automated tool for literature review writing.

> *Bachmann et al. (2024) explore the role of "brokerage" in facilitating new collaborations and its impact on academic success. Duan et al. (2024) analyze the importance of postdoctoral training for early-career success in academia. Kim et al. (2022) propose new multi-agent reinforcement learning challenges in StarCraft to explore the capabilities of AI algorithms in complex, multi-stage tasks. – from LR 9A*

**4.4 Woody Transitions**

The lack of stylistic variation in transitions is another notable shortcoming. Lots of the AI-generated reviews adopted predictable and repetitive transitional phrases, which might negatively impact the flow and readability of the text. For example, some reviews frequently use formulaic transitions like "several studies highlight," or "similarly," creating a monotonous tone that contrasts sharply with the nuanced and dynamic transitions typically employed by human writers. This stylistic limitation further reduced the overall quality and engagement of the reviews.

**4.5 Inconsistent Use of Tense**

A frequent grammatical issue in the generated reviews is the inconsistent use of tense when describing previous research. For instance, the reviews often mix past and present tenses without a clear rationale, sometimes within the same paragraph. This inconsistency creates a sense of randomness and diminishes the professionalism of the text.

> *Daskalaki et al. (2024) found that while educators recognize the potential of AI, they also express concerns about its impact on student privacy, particularly regarding the collection and use of student data. Similarly, Adanyin (2024) reveals consumer anxieties about data collection and management by AI systems in retail, highlighting a broader societal concern about data privacy that extends to the educational context. – from LR 9B*

### 4.6 Inclusion of Irrelevant Content

Even though this is not a common issue, it is still worth noticing that a few AI-generated literature reviews include some irrelevant content, often stemming from hallucination or misinterpretation of the context. These content, while related to broader AI topics, usually did not have direct relations to the specific focus of the review. Therefore, these irrelevant sections might mislead readers or detract from the focus of the review. In general, this issue could make people challenge the reliability of AI-generated outputs.

> *Additional research explores broader contexts and applications of AI that can inform its integration in education. Gavade et al. (2023) studied sociotechnical support infrastructures for teachers, highlighting the importance of support systems for teacher well-being, a factor that should be considered when implementing AI tools that might impact teacher roles and workload. – from LR 1A (topic: AI reduces teacher workload by automated grading and feedback)*
> *Beatty (2005) explored the use of classroom communication systems (CCSs) to enhance interactive pedagogy, demonstrating how technology can fundamentally transform the learning process and promote active student engagement. – from LR 8B (topic: AI assists teachers in curriculum development and classroom management)*

## 5. Discussions

The findings underscore significant challenges in using LLMs as a fully automated tool for generating literature review for academic purposes, particularly regarding their tendency to prioritise length and breadth over quality and depth. These issues align with and expand upon insights from prior research, offering a nuanced perspective on the current limitations of LLMs.

### 5.1 Repetition and the Impact of Context Window Size

The repetition observed in reviews is consistent with the observations of Shinde, Roy, and Ghosal (2022). Their work highlights that while LLMs can integrate longer references with clearer prompts and more inputs, the models with larger input sizes tended to create more content redundancy due to the model's inability to effectively prioritise and structure information. This is further supported by Yun et al. (2023), who found that LLMs sometimes "circle back" to previously generated content within a single systematic literature review. Our findings add to this body of knowledge by identifying the direct relationship between larger context windows and a higher frequency of redundant content.

In contrast, smaller context windows, as evidenced in Category B reviews, mitigate the repetition issue by forcing the model to work within a more confined scope. However, this trade-off limits the comprehensiveness of the review, as smaller context windows can result in fragmented narratives and a lack of necessary details of the cited works. This aligns with findings by Liu, Peng, and Weng (2023), who observed that models with smaller context windows like ChatGPT often generate superficial summaries of topics or studies in medical fields, showcasing great limitations in extracting and synthesising the key information in research.

### 5.2 Descriptive Nature vs. Analytical Depth

The descriptive nature of AI-generated literature reviews reflects another broader limitation of LLMs (Hsu et al., 2024; Yun et al., 2023). The "piling up" of superficial summaries of research and the tendency of prioritising content length over critical insights we observed particularly in reviews generated with large context window LLMs contribute another proof of this limitation. While most LLMs, especially after prompting and tuning, excel at summarising individual papers (Wang & Luo, 2024), they encounter difficulties in synthesising multiple sources into coherent analytical frameworks.
This descriptive tendency is compounded by the models' lack of domain-specific expertise, as noted by Kim et al. (2024). While human researchers rely on deep subject knowledge to identify patterns and contradictions within the literature, LLMs failed to offer the same level of analysis and interpretation of the results of the research. This finding highlights the necessity of human intervention to provide the analytical depth necessary for academic publication.

### 5.3 First Draft Helper

The findings above reinforce the point that AI-generated reviews should serve as foundational tools. The current LLMs are far from being a stand-alone tool for fully automating literature reviews. The omission of influential studies in AI-generated reviews is another significant challenge that needs to be solved. Although we acknowledge that our limited archive choices may have influenced the occurrence of the issue, our findings are consistent with recent studies. Antu, Chen, and Richards (2023) emphasised that LLMs sometimes fail to prioritise high-citation or seminal works, especially when the authors have disadvantaged backgrounds. Thus, we argue that the limitation stems from LLMs' overreliance on training data and the lack of ability to independently discern the relative importance of cited studies.
Furthermore, the omission of key works mirrors broader concerns about the "black box" nature of LLMs (Hassija et al., 2024), which can obscure the rationale behind citation choices. Unlike human scholars, who consciously choose sources to reflect the most influential contributions to a field, LLMs struggle to have contextual understanding required to perform this task effectively.

However, it is equally important to remember that all the AI-generated reviews, though not perfect, received at least a good mark from human experts. This finding suggests that LLMs have the potential to reduce the time and effort in writing a literature review. With some human-in-the-loop scaffoldings and revisions, it could be a valuable tool for drafting (Yun et al., 2023).

### 5.4 Stylistic and Grammatical Challenges

The importance of human editing and reviewing is underscored by the stylistic limitations observed in AI-generated reviews, including repetitive transitions and inconsistent tense usage. However, further research is needed to examine whether this issue is related to the language used in our prompts as the formality of the prompting might have some influence on the readability of the AI-generated outputs (Rawte et al., 2023). Moreover, even though some human authors tend to employ varied and nuanced transitions to maintain reader engagement and use tense in a consistent way, it is crucial to keep in mind that human writers are also not perfect (Uittenhove et al., 2024) and can write inconsistent sentences. Thus, we emphasise the need for human-AI partnership in the literature review writing process, a view echoed by many scholars.

### 5.5 The Role of Human Expertise in AI-Generated Reviews

The limitations identified in this study reinforce the findings of Yun et al. (2023), who argue that AI-powered tools are most effective when used as assistive technologies rather than standalone solutions. By integrating human expertise into the review process, researchers can address issues such as redundancy, omissions, and stylistic inconsistencies, ensuring that the final output meets the standards of academic rigor. This human-in-the-loop approach is particularly critical for non-expert users, who may lack the domain knowledge required to critically evaluate AI-generated outputs. As Antu et al. (2023) suggest, the role of human oversight extends beyond simple error correction; it involves actively curating and synthesising the literature to produce a cohesive and insightful narrative. Our findings support this view, emphasizing that while LLMs can streamline the initial stages of literature review generation (Uittenhove et al., 2024), they cannot replace the nuanced judgment and critical thinking of human researchers.

## 6. Limitations and Future Work

This research is a first step in exploring the efficacy of large language models in generating academic literature reviews. This work is still in the early stages of development and has some limitations that can be addressed in future.

All the AI generated literature reviews used in this research were generated by Gemini 1.5 pro. A number of other states of the art, closed-source and open-source large language models from OpenAI, Anthropic, Google, Meta and other AI labs can be used and their efficacy explored in the future. It would also be worth exploring the effectiveness of a large language model specifically fine-tuned on academic corpus of a few million open access papers.

The literature reviews generated and evaluated in this research were specifically on AI and education and reviewed by two domain experts and researchers. The quality of reviews generated in different domains may vary depending on the quality of open access publications and LLMs analysis. Changing the academic databases from which papers are extracted, quality of search implemented by those academic databases via Api and the number of relevant papers extracted can also impact the quality and length of literature reviews. Future research can focus on evaluating the AI generated literature reviews in diverse domains, evaluated by various researchers and domain experts.

## 7. Conclusion

This research is a first step in examining the capabilities and limitations of AI-generated literature reviews using LLMs with varying context window sizes. Our findings demonstrate that while LLMs with larger context windows offer significant advantages in incorporating broader information and maintaining coherence across extended inputs, they also introduce challenges, including content repetition, omission of critical works, and a lack of analytical synthesis. Reviews generated by smaller context windows mitigate redundancy but often lack the comprehensiveness necessary for academic rigor.

The descriptive nature of AI-generated reviews, coupled with stylistic limitations such as repetitive transitions and inconsistent tense usage, underscores the need for human oversight. The omission of seminal works and the inclusion of irrelevant content further highlight the limitations of LLMs in autonomously producing high-quality academic outputs. Despite these shortcomings, all reviews in this study achieved at least a "good" rating, suggesting that AI can serve as a valuable tool for drafting initial literature reviews, saving time and effort for researchers.

The findings emphasize the critical role of human expertise in refining AI-generated outputs. Human-in-the-loop approaches are essential to address issues such as redundancy, stylistic inconsistencies, and gaps in content, ensuring the final outputs meet the standards of academic scholarship. This is particularly vital for non-expert users who may lack the domain knowledge required to evaluate AI-generated literature critically.

Future research should explore the integration of advanced LLM architectures and fine-tuned models, specifically trained on domain-specific academic corpus, to enhance the quality of AI-generated outputs. Additionally, examining the impact of different academic databases, hybrid approaches, and user-friendly AI tools such as ResearchPal can further optimize the literature review process. While LLMs hold immense potential, their role should complement, rather than replace, human judgment and critical thinking in academic research.

## 8. Impact Statement

This study underscores the transformative potential of large language models (LLMs) in academic workflows, particularly for generating literature reviews. By comparing outputs from models with short and long context windows, the research highlights significant advantages in handling extensive information and maintaining coherence. However, it also reveals critical limitations, including redundancy, lack of analytical depth, and omissions of key works, emphasizing the necessity for human oversight. These findings have profound implications for the integration of AI in academic research, advocating for a hybrid approach that combines AI efficiency with human expertise to ensure rigor and reliability. Future advancements in LLMs and domain-specific fine-tuning hold the promise of enhancing the role of AI in academia while addressing current shortcomings. We have made all the code and data used in this paper public on a Github repo[2].

---

2 https://github.com/VeraciousAI/Long_Context_Paper

## Appendix

| Rubric | Rating | | | | | Score |
|---|---|---|---|---|---|---|
| | **5** | **4** | **3** | **2** | **1** | |
| Organization:<br><br>An organized and logical structure with transitions between sections and ideas | Organizing the review in a logical manner facilitates the flow of information for the reader | A logical organization and clear transitions are present in the review | In only a portion of the review is the content logically organized; the connections between sections or the ideas are weak | There is a lack of organization in the review and no connection between sections and ideas | The review is not organized in a logical manne | |
| Introduction: Includes background, significance, and good structure | All elements of the Introduction present, leading to a strong, structure | All elements of the Introduction present, leading to a logical structure | 1 element of the introduction is missing | 2 of the 3 elements of the introduction are missing | All important elements of the introduction are missing | |
| Synthesis: Articles are used to support ideas, rather than just descriptive summariesarized | Throughout the paragraph, studies are used to support the main idea of the paragraph, with strong connections between them | Good connections made between studies. Studies generally used to support ideas | Studies are interconnected. however, in some parts, only summaries of studies are included | There are some weak connections between studies, but mostly summaries | The studies have no connection. Studies are merely summarized | |
| Analysis: critical thinking is present and, evaluation of studies are indicated | Strong critical evaluation of the literature is demonstrated | Thee evaluation of the literature is appropriately integrated | The evaluation of the literature is presented, but not always integrated | Some evaluation of the literature is presented, but not well integrated | No evaluation of the literature | |
| Conclusion: Synthesis of all ideas with identifying a clear research gap) | Synthesis of all ideas and a satisfying closing | Synthesis of most ideas, logically concludes the review | Generally addresses most ideas and logically concludes the review | Some ideas addressed, and review concluded | No conclusion | |
| Content accuracy | content is 100% accurate | content is 90% accurate | content is 80% accurate | content is 70% accurate | content is 60% accurate | |
| Relevancy of the content | content is 100% relevant | content is > = 90% relevant | content is > = 80% relevant | content is > = 70 and > 60 % relevant | content is < = 60% relevant | |
| Objectivity (Ensures that a acadimic tone is used) | acadimic tonethroughout | Tone is generally acadimic tone with one or two lapses. | Tone is generally acadimic tone, with some lapses | Tone is acadimic tone some of the time, though often not | Completely Informal | |
| **Citations and References** | | | | | | |
| Currency of the citation | Majority recent 5 years | Majority Recent 10 years | Majority Recent 15 years | Majority Recent 20 years | Majority Recent 50 years | |

| Format of the reference List | There is proper citation of information and the format is in accordance with APA standards. | Information is cited in APA format, but has errors.<br>Or not recent version | IAlthough the information is cited, it is not formatted in the APA style | It is not possible to retrieve or assign the text to the references due to insufficient information in the references | INo citations are provided for the informationI | |
|---|---|---|---|---|---|---|
| Using and formatting references | The citation of information is consistent, accurate, and formatted correctly | IThere is a consistent citation of information, but the format is incorrect | There are errors in the citation and the format is incorrect | Citations are inconsistent, and the format is incorrect | Citation not provided | |
| Type of references | Information presented comes from at least 100% separate, peer-reviewed, research sources | Information presented comes from at least >= 80% peer-reviewed, research source | Information presented comes from at least >= 60 peer-reviewed peer-reviewed, research sources | Information presented comes from at least >= 40% or fewer peer-reviewed, research sources | Information presented comes from unverified sources | |
| | | | **Presentation** | | | |
| Grammar | No more than 2 grammar errors | Few Grammar errors | Reader is annoyed by grammar errors, but they don't interfere with the reading. | Too many grammatical errors | Riddled with grammar errors | |
| Spelling and punctuation | There are no errors | < 3 errors | < 5 errors | < 10 errors | ≥ 10 errors | |
| Conjunctions | Varieties of conjunctions types of conjunctions 7 or more conjunction words | some variation of conjunctions using less than 7 conjunction words | Same types of conjunctionsimilar 5 or less conjunction words | 3 conjunctions | 0-2 conjunctions | |
| **Total Points** | | | | | | |